\documentclass[runningheads]{llncs}

\usepackage[T1]{fontenc}
\def\doi#1{\href{https://doi.org/\detokenize{#1}}{\url{https://doi.org/\detokenize{#1}}}}
\usepackage{booktabs}
\usepackage{multirow}
\usepackage{graphicx, color, soul}

\usepackage[misc,geometry]{ifsym}
\usepackage{hyperref}
\usepackage{listings}
\begin{document}
\title{Automatic fetal fat quantification from MRI}
%
%

\author{Netanell Avisdris* \textsuperscript{\Letter} \inst{1,2} \orcidID{0000-0002-0719-1327} \and  
Aviad Rabinowich*\inst{2,3,4} \and  
Daniel Fridkin\inst{1} \and  
Ayala Zilberman\inst{4,5} \and  
Sapir Lazar\inst{3,4} \and  
Jacky Herzlich\inst{4,6} \and  
Zeev Hananis\inst{2} \and 
Daphna Link-Sourani\inst{4} \and 
Liat Ben-Sira\inst{3,4} \and
Liran Hiersch\inst{4,5} \and 
Dafna Ben Bashat\inst{2,4,7} \and 
Leo Joskowicz\inst{1} 
}

%
\authorrunning{N. Avisdris, A. Rabinowich et al.}
%
\institute{School of Computer Science and Engineering, The Hebrew University of Jerusalem, Israel \email{\{netana03,josko\}@cs.huji.ac.il} \and
Sagol Brain Institute, Tel Aviv Medical Center, Israel \and Department of Radiology, Tel Aviv Medical Center, Tel Aviv, Israel\and Sackler Faculty of Medicine, Tel Aviv University, Tel Aviv, Israel\and Department of Obstetrics and Gynecology, Lis Hospital for Women, Tel Aviv  Sourasky Medical Center, Tel Aviv, Israel\and Neonatal Intensive Care Unit, Dana Dwek Children's Hospital, Tel Aviv Sourasky Medical Center, Tel Aviv, Israel\and
Sagol School of Neuroscience, Tel Aviv University, Tel Aviv, Israel 
}

\maketitle 
\begin{abstract}
Normal fetal adipose tissue (AT) development is essential for perinatal well-being. AT, or simply fat, stores energy in the form of lipids. Malnourishment may result in excessive or depleted adiposity. Although previous studies showed a correlation between the amount of AT and perinatal outcome, prenatal assessment of AT is limited by lacking quantitative methods. Using magnetic resonance imaging (MRI), 3D fat- and water-only images of the entire fetus can be obtained from two-point Dixon images to enable AT lipid quantification. This paper is the first to present a methodology for developing a deep learning (DL) based method for fetal fat segmentation based on Dixon MRI. It optimizes radiologists' manual fetal fat delineation time to produce annotated training dataset. It consists of two steps: 1) model-based semi-automatic fetal fat segmentations, reviewed and corrected by a radiologist; 2) automatic fetal fat segmentation using DL networks trained on the resulting annotated dataset. Segmentation of 51 fetuses was performed with the semi-automatic method. Three DL networks were trained. We show a significant improvement in segmentation times (3:38 hours $\rightarrow$ < 1 hour) and observer variability (Dice of 0.738 $\rightarrow$ 0.906) compared to manual segmentation. Automatic segmentation of 24 test cases with the 3D Residual U-Net, nn-UNet and SWIN-UNetR transformer networks yields a mean Dice score of 0.863, 0.787 and 0.856, respectively. These results are better than the manual observer variability, and comparable to automatic adult and pediatric fat segmentation. A radiologist reviewed and corrected six new independent cases segmented using the best performing network (3D Residual U-Net), resulting in a Dice score of 0.961 and a significantly reduced correction time of 15:20 minutes. Using these novel segmentation methods and short MRI acquisition time, whole body subcutaneous lipids can be quantified for individual fetuses in the clinic and large-cohort research.
\end{abstract}
\section{Introduction}
The adipose tissue (AT) is essential for fetal development and reflects the overall fetal energy balance. AT, or simply fat, stores energy in the form of lipids. Typically, well-nourished fetuses show an accelerated AT growth from the 28 weeks of gestation onward~\cite{lee2005fetal}. 
Although birth weight is often used as a proxy of fetal nutrition and as a predictor of adverse perinatal outcome, prior studies suggest that fetal body fat may show high correlation and be better predictor for neonatal outcomes
~\cite{banting2022estimation,carberry2013body,shaw2019does}.
Previous studies using ultrasound (US) showed alternations of fetal AT related to fetal growth restriction and excess of fetal AT in cases of maternal diabetes \cite{gardeil1999subcutaneous,larciprete2005intrauterine,nobile2010growth}. Furthermore, these changes also correlate with neonatal outcomes, emphasizing the clinical relevance of AT quantification. Although US is the method of choice for fetal development assessment, it is limited by the lack of true 3D information. Thus, fat estimation currently relies on linear and area measurements of selected fetal body parts, e.g., the abdomen and the limbs, and on estimated fractional limb volume \cite{lee2009fracvol}, with no full body AT fat content volume quantification and analysis. 

Magnetic resonance imaging (MRI) provides 3D multi-contrast information that indirectly characterizes the microstructural properties of tissue. For fat assessment, the two-point Dixon method \cite{dixon1984simple}, a proton chemical shift MRI technique that produces separated fat-only and water-only images from a dual-echo acquisition, is used. In this method, water and fat signals alternate between being summed and subtracted, yielding fat-only and water-only images that can be analyzed to quantify lipids. This method is used to quantify the lipid content of various organs, most extensively for hepatosteatosis assessment \cite{hu2021linearity}.

Fetal MRI is increasingly used as a complementary method to US, mainly for detecting central nervous system (CNS) and non-CNS anomalies, including thoracic, gastrointestinal, genitourinary, and skeletal anomalies \cite{cassart2020european}. A few studies measured AT of fetuses with normal-growth and with maternal diabetes \cite{berger2012quantification,blondiaux2018developmental,giza2021water}. However, these studies are limited, as they rely on linear measurements \cite{berger2012quantification}, on local assessment \cite{blondiaux2018developmental}, or require laborious manual segmentation \cite{giza2021water}.

Automatic 3D MRI segmentation can enable accurate and reliable routine AT lipid quantification. Recently, deep learning (DL) based models have been increasingly used for automated segmentation of structures in medical imaging, including fetal structures \cite{meshaka2022artificial,torrents2019fetalsegmentationreview}. While most methods address fetal brain segmentation, a few have been developed for other fetal structures and organs.

To the best of our knowledge, no automatic or semi-automatic methods for fetal fat quantification in US or MRI have been developed. Roelants et al. \cite{roelants2017foetal} described automated methods for US limb soft-tissue quantification, a surrogate for subcutaneous AT lipid deposition. Mack et al. \cite{mack2016semiautomated} presented a semi-automatic method for fractional limb volume assessment and showed that it reduces observer variability and annotation time. Recent papers described automatic DL methods for fat segmentation in Dixon scans of adults in the pancreas \cite{lin2022pancreas_seg_adult} and 
visceral and subcutaneous fat \cite{estrada2020fatsegnet,kway2021automated}. Estrada et al. \cite{estrada2020fatsegnet} reported that manual fat delineation is tedious and time-consuming, thus limiting its clinical and research applications. 
Fat segmentation poses several significant challenges compared to other structures. First, ground-truth manual fetal fat delineations are time-consuming, laborious and difficult to acquire \cite{estrada2020fatsegnet,giza2021water}, and 
suffer from high observer variability \cite{joskowicz2019observervar}. Indeed, fat structure is thin and complex, is distributed in various locations, and may appear as multiple disjoint components with a wide variety of shapes. Moreover, Dixon scans exhibit inherent limitations, e.g., signal inhomogeneity, artifacts due to high sensitivity to fetal and maternal motion, and obscured lipid-poor tissues in fat-only images, which lack the fetus context without the water-only images and requires additional MRI sequences.

In this paper, we present a methodology for creating the first reported automatic DL-based method for fetal fat segmentation on Dixon images. Its contributions are three-fold: (1) a semi-automatic segmentation method for fetal fat delineation that substantially shortens the manual annotation time and reduces inter-observer variability; (2) training and evaluation of three state-of-the-art deep learning models for fetal fat segmentation on the validated ground-truth segmentations generated by the semi-automatic method; (3) quantification of the manual and semi-automatic delineation observer variability and annotation time of fetal fat. 

\section{Methodology}

Our methodology for fetal fat segmentation consists of: 1) semi-automatic fetal fat model-based segmentation whose aim is to shorten the manual fetal fat delineation time required to produce annotated training data, and; 2) automatic fetal fat segmentation with DL networks trained on the resulting annotated dataset.

\subsection{Semi-automatic fetal AT segmentation}
\begin{figure}[t]
\includegraphics[width=\textwidth]{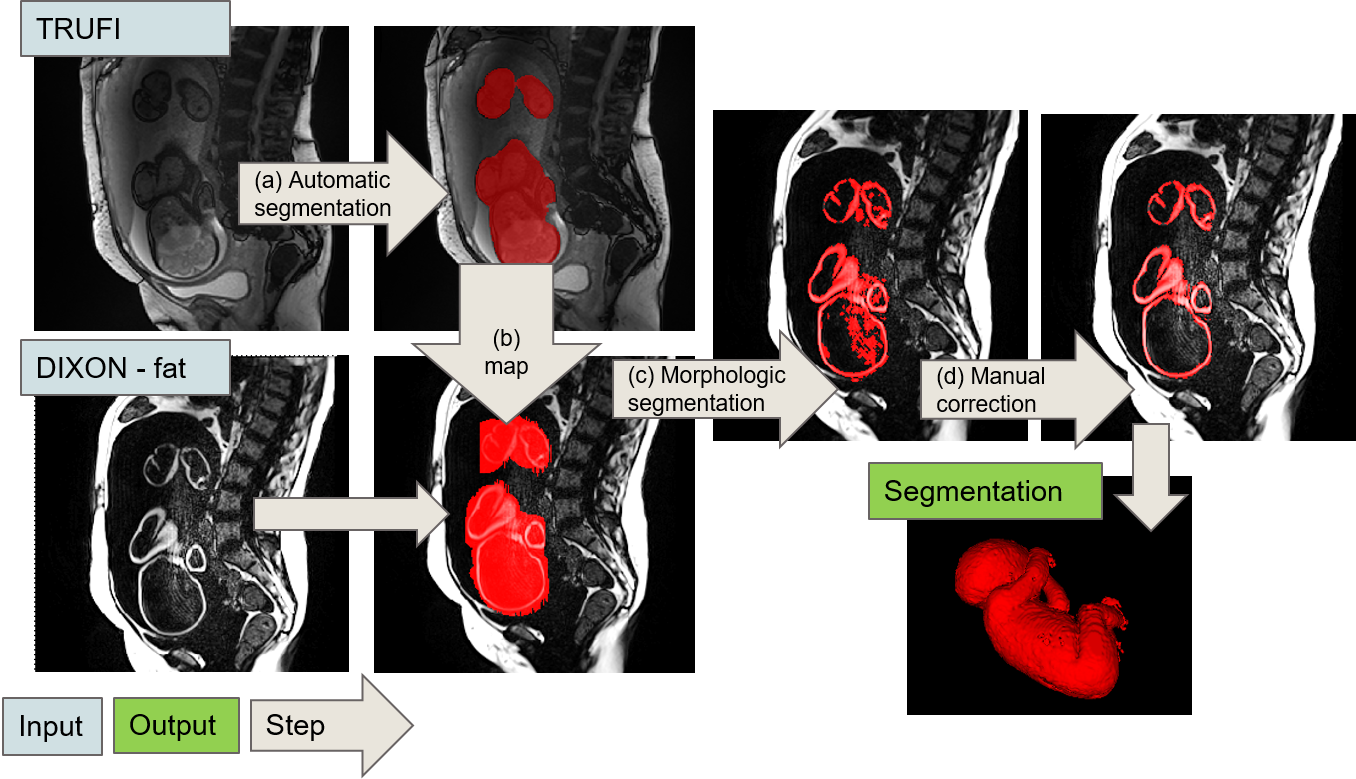}
\caption{Semi-automatic fetal fat segmentation method. The inputs (light blue) are fetal body TRUFI and fat-only Dixon scans. The four steps (grey arrows) of the method are: (a) automatic segmentation of the fetal body in the TRUFI scans; (b) mapping of the resulting segmentation as a VOI on the fat-only Dixon volume; (c) automatic segmentation of the fetal fat on the Dixon scan by thresholding and morphology operations; (d) revision and manual correction by a radiologist of the resulting fetal fat segmentation mask on each slice. The output (green) is a validated fetal fat segmentation mask. \looseness = -1} \label{fig_semi}
\end{figure}
The inputs for the semi-automatic fetal fat segmentation method are fetal body TRUFI and fat-only Dixon MRI sequences. The output is an initial subcutaneous fetal fat segmentation on the Dixon scan (Fig.~\ref{fig_semi}). The TRUFI sequence is used to obtain fetal body segmentation that defines the Volume of Interest (VOI) in the Dixon scan on which the fetal fat segmentation is computed. Although the TRUFI and Dixon scans are acquired at subsequent times, they have different field-of-views and may not be aligned due to fetal and maternal motion. 

First, the fetal body is automatically segmented on the TRUFI scan with the DL model described in \cite{dudovitch2020deep}. The resulting segmentation mask is then mapped to the Dixon scan using the scanning position information to define a VOI that includes the entire fetus and excludes the maternal abdominal fat regions. Manual adjustments to the resulting VOI are performed as required. The fetal fat within the VOI is then thresholded with a pre-defined value chosen experimentally. Connected components with < 50 voxels are discarded to remove artifacts with similar fat intensity. The result is then reviewed and corrected by an expert radiologist to produce high-quality, validated ground truth segmentation masks. 

\subsection{Automatic fetal fat segmentation}
\begin{figure}[tp]
\includegraphics[width=0.8\textwidth]{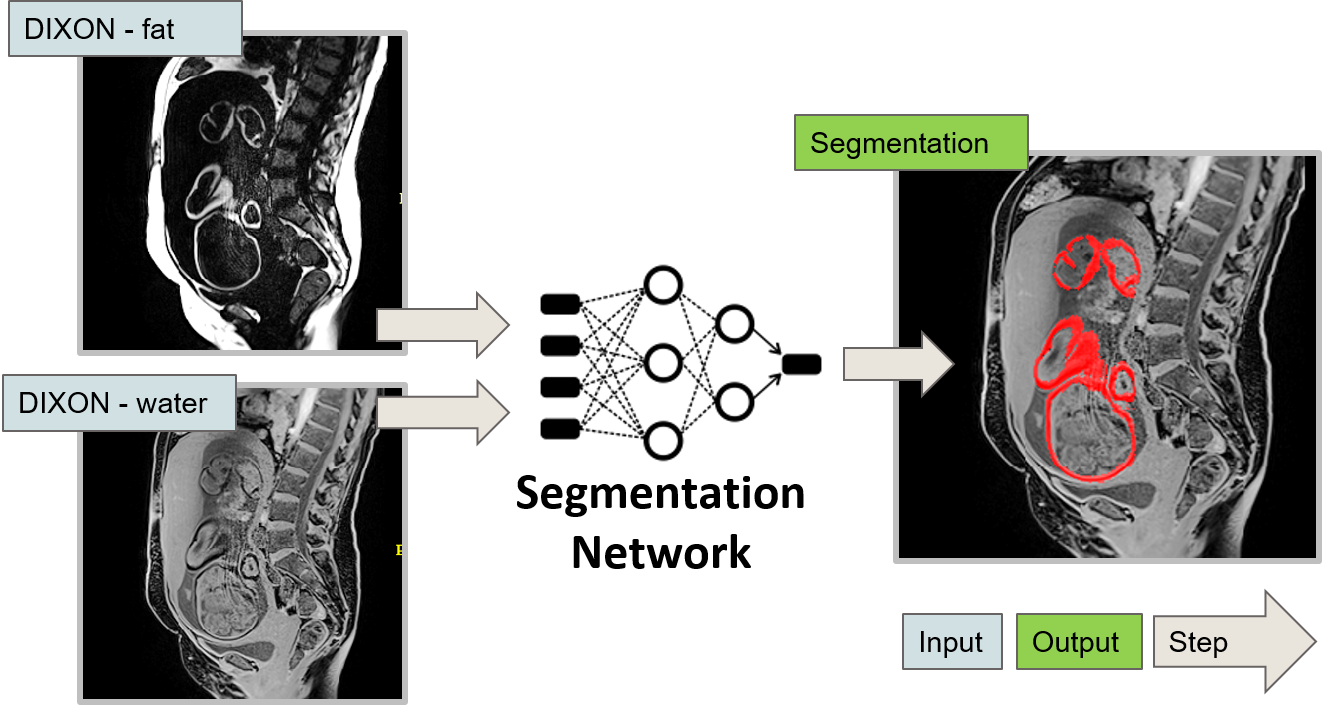}
\centering
\caption{Automatic fetal fat segmentation. The inputs (light blue) are fat-only and water-only Dixon scans. The output (green) is the fetal fat segmentation with a DL network. \looseness = -1} \label{fig_autoseg}
\end{figure}
Automatic fetal fat segmentation is performed with a state-of-the-art DL model trained on the high-quality annotations of the fetal fat created in the previous step. The inputs are fat-only and water-only Dixon volumes, which are acquired simultaneously, and are thus aligned. The output is the fetal fat segmentation. 

We evaluate three models: (1) Residual 3D U-Net \cite{kerfoot2018resunet}; (2) nn-UNet \cite{isensee2021nnunet}, and; (3) SWIN-UNetR \cite{tang2022swin}. The 3D U-Net and nn-UNet are fully convolutional (FCN) encoder-decoder networks, where the decoder network is connected to the encoder network through skip connections. SWIN-UNetR is a combined transformer-FCN network. We briefly describe them next. 

\textbf{Residual 3D U-Net} is a 3D-UNet \cite{cciccek20163dunet} in which the encoder and decoder network blocks are residual units. The 3D U-Net is a FCN which consists of a contraction and an expansion path. The contraction path is an encoder that captures the context, while the expansion path is a decoder network that performs up-sampling to recover the segmentation map size. The encoder and decoder paths are connected by path skip connections to share localization information.

\textbf{nn-UNet} is a biomedical image segmentation framework that automatically adapts to a dataset characteristics. It selects a segmentation network from a wide range of options, including 2D- and 3D- U-Net, and configures their hyper-parameters, e.g., patch size, batch size, or learning rate. It achieves state-of-the-art results on a wide range of biomedical image segmentation scenarios.

\textbf{SWIN-UNetR}: Transformers are new DL architectures that achieved state-of-the-art results in a wide range of machine learning tasks, including medical imaging \cite{shamshad2022transformers}. Shifted windows (SWIN) Transformers is a hierarchical vision transformer that allows for local computing of self-attention with non-overlapping windows. SWIN is more efficient than regular vision transformers and is well-suited for tasks requiring multi-scale modeling due to its hierarchical nature.is a vision transformer for biomedical image analysis. It consists of a SWIN transformer as the encoder and a CNN-based decoder. SWIN-UNetR computes self-attention in an efficient shifted window partitioning scheme. It is currently the best performing model for a wide variety of biomedical image segmentation tasks.

\textbf{Implementation details:} The 3D-UNet and SWIN-UNetR models were implemented in MONAI~\cite{monai_consortium_2022_6639453}; nn-UNet was implemented in Pytorch. All networks were trained on 128x128x128 patches for 300 epochs on a single NVIDIA V100 GPU. The 3D-UNet was trained with an ADAM optimizer with an initial learning rate of $1 \times 10^{-2}$ and batch size of 10. SWIN-UNetR was trained with an AdamW optimizer with a warm-up cosine scheduler of 50 iterations and batch size of 1 patch. Both 3D-UNet and SWIN-UNetR are trained with random patch cropping and Gaussian noise augmentations, with Dice loss. Since nn-UNet is an automated framework, no training hyper-parameters were selected. For inference, we use a sliding window with an overlap of 0.7 for neighboring voxels.

\section{Experimental Results}
We conducted three studies. Study 1 quantifies the observer variability of manual and semi-automatic fat delineation. The manual delineation observer variability provides a reference for the expected target accuracy; the semi-automatic delineation observer variability quantifies the expected improvement in observer agreement. The study also measures the radiologist time required for each method. Study 2 quantifies the performance of the three automatic DL fetal fat segmentation methods. Study 3 quantifies and analyze the segmentation accuracy and the correction time using the best-performing automatic DL network.

\textbf{Study population:}
Retrospective fetal MRI studies were collected at the Tel Aviv Sourasky Medical Center, Israel, 
between 2019 and 2022. The Institutional Review Board approved the study and waived the need for informed consent. The dataset consisted of 57 singleton fetuses ranging between 31 and 39+1 gestation weeks (mean = 33.9, std = 1.8). Participants were referred to fetal MRI for various clinical indications. Cases with chromosomal or congenital anomalies were excluded. 

\textbf{Data acquisition and ground truth generation:}
Patients were scanned on one of three 3T MRI scanners (Skyra, Prisma and Vida, Siemens Healthineers). Two subsequent sequences were used for this study: (a) free-breathing T2-weighted TRUFI sequence with voxel resolution of $0.78 \times 0.78 \times 2 mm^3$ and an acquisition time of ~60 seconds, and (b) two-point Dixon sequence with voxel resolution of $1.25-1.4 \times 1.25-1.4 \times 1.5-2mm^3$. The Dixon sequence was acquired with a single breath-hold, with acquisition time of 18-20 seconds. ITK-SNAP(v. 3.8.0) \cite{yushkevich2016itk} was used for manual delineation and segmentation correction.

\textbf{Evaluation metrics}: Five metrics were used to estimate observer agreement and evaluate segmentation methods: Dice similarity coefficient, Hausdorff distance, Average Symmetric Surface Distance (ASSD), volume difference (VD), and relative volume difference (RVD). RVD is defined as the VD divided by the fetal body volume. We used a two-sided t-test to estimate statistical significance; $p < 0.05$ was considered significant.

\begin{table}[t]
\resizebox{\columnwidth}{!}{%
\begin{tabular}{l|cccc||cccc|}
\cline{2-9}
 &
  \multicolumn{4}{c||}{Manual} &
  \multicolumn{4}{c|}{Semi-automatic} \\ \cline{2-9} 
 &
  \multicolumn{1}{c|}{mean} &
  \multicolumn{1}{c|}{std} &
  \multicolumn{1}{c|}{min} &
  max &
  \multicolumn{1}{c|}{mean} &
  \multicolumn{1}{c|}{std} &
  \multicolumn{1}{c|}{min} &
  max \\ \hline
\multicolumn{1}{|l|}{Dice} &
  \multicolumn{1}{c|}{0.738} &
  \multicolumn{1}{c|}{0.092} &
  \multicolumn{1}{c|}{0.567} &
  0.865 &
  \multicolumn{1}{c|}{\textbf{0.906}} &
  \multicolumn{1}{c|}{\textbf{0.084}} &
  \multicolumn{1}{c|}{0.744} &
  0.981 \\ \hline
\multicolumn{1}{|l|}{Hausdorff {[}mm{]}} &
  \multicolumn{1}{c|}{21.09} &
  \multicolumn{1}{c|}{24.66} &
  \multicolumn{1}{c|}{2.24} &
  86.82 &
  \multicolumn{1}{c|}{\textbf{16.88}} &
  \multicolumn{1}{c|}{\textbf{5.10}} &
  \multicolumn{1}{c|}{10.44} &
  28.32 \\ \hline
\multicolumn{1}{|l|}{ASSD {[}mm{]}} &
  \multicolumn{1}{c|}{1.90} &
  \multicolumn{1}{c|}{4.11} &
  \multicolumn{1}{c|}{0.23} &
  13.57 &
  \multicolumn{1}{c|}{\textbf{0.36}} &
  \multicolumn{1}{c|}{\textbf{0.28}} &
  \multicolumn{1}{c|}{0.07} &
  0.87 \\ \hline
\multicolumn{1}{|l|}{VD {[}mL{]}} &
  \multicolumn{1}{c|}{-} &
  \multicolumn{1}{c|}{-} &
  \multicolumn{1}{c|}{-} &
  - &
  \multicolumn{1}{c|}{18.69} &
  \multicolumn{1}{c|}{20.63} &
  \multicolumn{1}{c|}{2.50} &
  73.48 \\ \hline
\multicolumn{1}{|l|}{RVD {[}\%{]}} &
  \multicolumn{1}{c|}{29.91} &
  \multicolumn{1}{c|}{12.54} &
  \multicolumn{1}{c|}{6.26} &
  46.34 &
  \multicolumn{1}{c|}{\textbf{9.26}} &
  \multicolumn{1}{c|}{\textbf{9.14}} &
  \multicolumn{1}{c|}{1.10} &
  26.50 \\ \hline 
\end{tabular}%
}\\[0.5ex]
\caption{Inter-observer variability of fetal fat delineation between two radiologists using two methods: manual and semi-automatic. Bold face indicates best results for each of the five metrics.} 
\label{tab:interob-tbl}
\end{table}

\subsection{Study 1: Manual and semi-automatic observer variability}
No previous study has analyzed the inter-observer variability in fetal fat segmentation. Manual and semi-automatic segmentations were assessed by two radiologists that were blinded to the study indication and to the gestational age. A previous study noted extremely long manual segmentation times \cite{giza2021water}. Therefore, four consecutive slices on 10 scans were randomly selected for manual segmentation in the fetal volume. For the semi-automatic segmentation, the first observer (R1) segmented the entire cohort. Ten cases were assigned to a second observer (R2) to establish a baseline for the automatic method and to assess the segmentation quality. Manual delineation and correction times were recorded.

Table~\ref{tab:interob-tbl} lists the results. Overall, manual fat segmentation was very time-consuming, with a mean of 3:38 hours (R1: 3:43 hours, R2: 3:34 hours) for the entire fat fetal volume, similar to the reported time in \cite{giza2021water}. Semi-automatic segmentation significantly reduced total segmentation time to a mean of 0:57 hours ($p=6.3 \times 10^{-7}$). Furthermore, semi-automatic delineation significantly increased the inter-observer agreement for Dice and RVD: Dice increased by 0.186 ($p=3 \times 10^{-5}$). Also, Hausdorff decreased by 4.20$mm$ ($p=0.585$), ASSD decreased by 1.54$mm$ ($p=0.228$) and RVD decreased by $20.55\%$ ($p=3 \times 10^{-5}$).

\subsection{Study 2: Automatic fetal AT segmentation}

\begin{figure}[htp]
\includegraphics[angle=90, width=\textwidth]{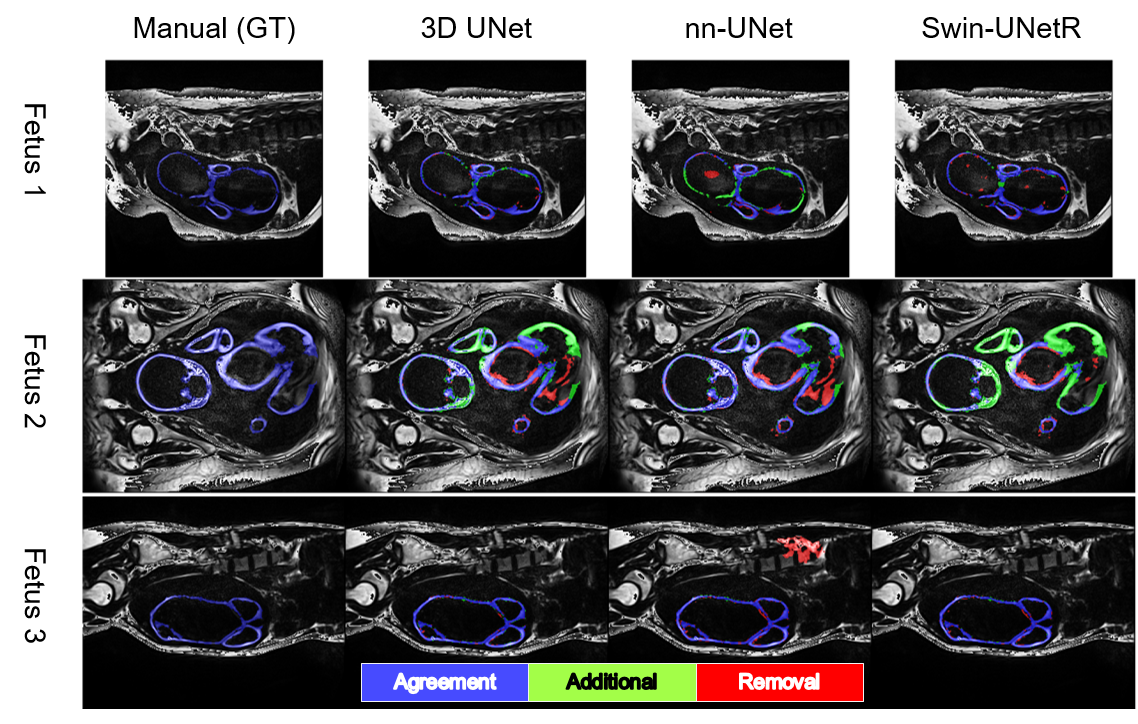}
\caption{Comparison of automatic segmentation networks performance. Three examples (columns) of manual ground-truth and three automatic segmentation networks (rows). The manual and automatic segmentation agreement voxels are shown in blue.
Under-segmentation is shown in green. Oversegmentation is shown in red. The nn-UNet tends to yield over-segmented components, including large out-of-fetus components (Fetus 3).   \looseness = -1} \label{fig_comap_results}
\end{figure}

We evaluate three networks for the automatic fetal AT segmentation task: 3D Residual U-Net, nn-UNet, and SWIN-UNetR. A dataset of 51 manually corrected volumes from semi-automatic fetal fat segmentation was used. The networks were trained on 21 volumes; six volumes were used for validation and hyper-parameter choice. The networks were tested on an independent set of 24 volumes. Results are listed on Table \ref{tab:automaticresults}. Overall, the 3D Residual UNet yielded the best results, achieving a mean Dice score of 0.862, above the manual delineation observer variability. Fig. \ref{fig_comap_results} shows illustrative fetal fat segmentation results of three fetuses.
\begin{table}[tp]
\resizebox{\columnwidth}{!}{%
\begin{tabular}{l|llll|llll|llll|}
\cline{2-13}
 &
  \multicolumn{4}{l|}{\textbf{3D Residual UNet}} &
  \multicolumn{4}{l|}{\textbf{nn-UNet}} &
  \multicolumn{4}{l|}{\textbf{SWIN-UNetR}} \\ \cline{2-13} 
 &
  \multicolumn{1}{c|}{\textbf{mean}} &
  \multicolumn{1}{c|}{\textbf{std}} &
  \multicolumn{1}{c|}{\textbf{min}} &
  \multicolumn{1}{c|}{\textbf{max}} &
  \multicolumn{1}{c|}{\textbf{mean}} &
  \multicolumn{1}{c|}{\textbf{std}} &
  \multicolumn{1}{c|}{\textbf{min}} &
  \multicolumn{1}{c|}{\textbf{max}} &
  \multicolumn{1}{c|}{\textbf{mean}} &
  \multicolumn{1}{c|}{\textbf{std}} &
  \multicolumn{1}{c|}{\textbf{min}} &
  \multicolumn{1}{c|}{\textbf{max}} \\ \hline
\multicolumn{1}{|l|}{\textbf{Dice}} &
  \multicolumn{1}{l|}{\textbf{0.862}} &
  \multicolumn{1}{l|}{\textbf{0.063}} &
  \multicolumn{1}{l|}{0.609} &
  0.926 &
  \multicolumn{1}{l|}{0.787} &
  \multicolumn{1}{l|}{0.169} &
  \multicolumn{1}{l|}{0.269} &
  0.927 &
  \multicolumn{1}{l|}{0.856} &
  \multicolumn{1}{l|}{0.082} &
  \multicolumn{1}{l|}{0.538} &
  0.933 \\ \hline
\multicolumn{1}{|l|}{\textbf{Hausdorff {[}mm{]}}} &
  \multicolumn{1}{l|}{\textbf{18.05}} &
  \multicolumn{1}{l|}{\textbf{4.88}} &
  \multicolumn{1}{l|}{9.70} &
  28.58 &
  \multicolumn{1}{l|}{37.19} &
  \multicolumn{1}{l|}{34.06} &
  \multicolumn{1}{l|}{9.43} &
  134.85 &
  \multicolumn{1}{l|}{20.63} &
  \multicolumn{1}{l|}{5.74} &
  \multicolumn{1}{l|}{9.17} &
  36.07 \\ \hline
\multicolumn{1}{|l|}{\textbf{ASSD {[}mm{]}}} &
  \multicolumn{1}{l|}{\textbf{0.60}} &
  \multicolumn{1}{l|}{\textbf{0.33}} &
  \multicolumn{1}{l|}{0.30} &
  1.93 &
  \multicolumn{1}{l|}{1.54} &
  \multicolumn{1}{l|}{2.09} &
  \multicolumn{1}{l|}{0.32} &
  7.52 &
  \multicolumn{1}{l|}{0.66} &
  \multicolumn{1}{l|}{0.51} &
  \multicolumn{1}{l|}{0.25} &
  2.78 \\ \hline
\multicolumn{1}{|l|}{\textbf{VD {[}mL{]}}} &
  \multicolumn{1}{l|}{\textbf{16.93}} &
  \multicolumn{1}{l|}{\textbf{27.16}} &
  \multicolumn{1}{l|}{0.74} &
  138.38 &
  \multicolumn{1}{l|}{27.67} &
  \multicolumn{1}{l|}{36.08} &
  \multicolumn{1}{l|}{1.53} &
  177.89 &
  \multicolumn{1}{l|}{19.45} &
  \multicolumn{1}{l|}{38.47} &
  \multicolumn{1}{l|}{0.19} &
  189.00 \\ \hline
\multicolumn{1}{|l|}{\textbf{RVD {[}\%{]}}} &
  \multicolumn{1}{l|}{\textbf{7.11}} &
  \multicolumn{1}{l|}{\textbf{7.44}} &
  \multicolumn{1}{l|}{0.50} &
  36.66 &
  \multicolumn{1}{l|}{13.59} &
  \multicolumn{1}{l|}{15.78} &
  \multicolumn{1}{l|}{0.82} &
  71.60 &
  \multicolumn{1}{l|}{8.36} &
  \multicolumn{1}{l|}{11.50} &
  \multicolumn{1}{l|}{0.12} &
  50.07 \\ \hline
\end{tabular}%
}\\[0.5ex]
\caption{Study 2 results: comparison of three networks for fetal AT segmentation: 3D Residual UNet, nn-UNet and SWIN-UNetR on test-set (n=24). Bold face indicates best results for a metric. }
\label{tab:automaticresults}
\end{table}

\subsection{Study 3: Analysis of manual corrections after automatic segmentation}
To evaluate the effectiveness of the automatic method, six additional test cases were segmented using the 3D Residual UNet, which was the best performing network. The resulting segmentations were manually corrected by a radiologist (R1). The manual correction times were recorded.
\begin{table}[htp]
\centering{
\resizebox{0.6\columnwidth}{!}{%
\begin{tabular}{l|l|l|l|l|}
\cline{2-5}
 & \multicolumn{1}{c|}{mean} & \multicolumn{1}{c|}{std} & \multicolumn{1}{c|}{min} & \multicolumn{1}{c|}{max} \\ \hline
\multicolumn{1}{|l|}{Dice}               & 0.961 & 0.025 & 0.915 & 0.983 \\ \hline
\multicolumn{1}{|l|}{Hausdorff {[}mm{]}} & 48.72 & 29.67 & 13.08 & 84.30 \\ \hline
\multicolumn{1}{|l|}{ASSD {[}mm{]}}      & 0.59  & 1.00  & 0.10  & 2.63  \\ \hline
\multicolumn{1}{|l|}{VD {[}mL{]}}        & 7.54  & 10.91 & 1.62  & 29.35 \\ \hline
\multicolumn{1}{|l|}{RVD {[}\%{]}}       & 4.68  & 5.48  & 1.37  & 15.56 \\ \hline
\end{tabular}%
}\\[0.5ex]
}
\caption{Study 3 results. Manual correction on best performing automatic segmentation network (3D Residual UNet) for 6 cases.}
\label{tab:correctioncost}
\end{table}

\begin{figure}[t]
\includegraphics[width=.9\textwidth]{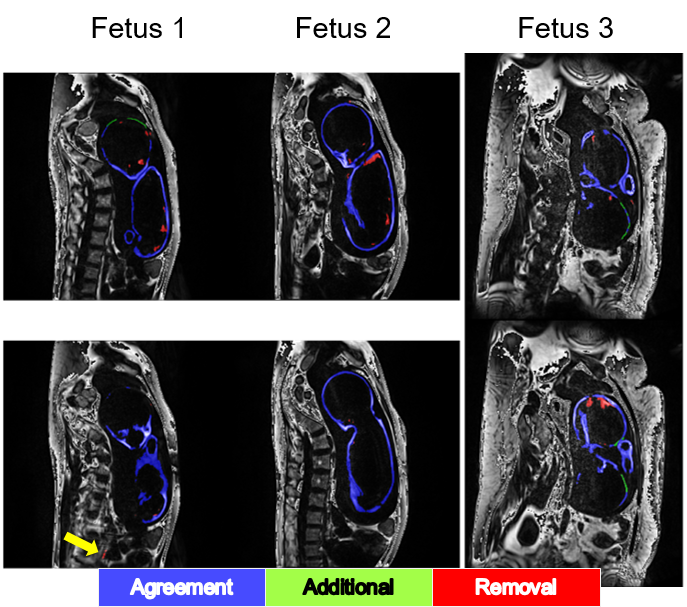}
\caption{Examples of manual correction on best performing automatic segmentation network (3D Residual UNet) of three fetuses (columns) from two representative slices (rows). The yellow arrow points to a small out-of-fetus segmented compartment that needs to be erased. \looseness = -1} \label{fig_corrections}
\end{figure}

Results are listed in Table \ref{tab:correctioncost}. Overall, only minor revisions were required (Dice of 0.961), and correction times were reduced to an average of 15:20 minutes. However, the Hausdorff distance was relatively high, 48.72mm, compared to the inter-observer one (16.88mm). Visual examples of corrections are presented in Figure \ref{fig_corrections}. Note that minor corrections were performed on small, out-of-fetus voxels (yellow arrow on figure \ref{fig_corrections}), which may explain this result. Additional corrections were performed on lipid-poor areas such as the scalp, intensity artifacts, and non-subcutaneous fat depots (e.g. , perirenal fat). All fetus body parts were segmented using the automatic method, with only minor voxels difference, resulting in a relatively small RVD of 4.68$\%$.

\section{Discussion}

In this paper, we describe methods for semi-automatic and automatic fetal subcutaneous fat segmentation. We propose a semi-automatic segmentation method to delineate the fetal subcutaneous fat and showed that it significantly improves segmentation times and observer variability compared to manual segmentation. We then used the resulting segmentation masks to train an automatic deep-learning network, which yields accurate results, better than the inter-observer variability on manual segmentation, and shorten the correction time.

Automatic fat segmentation, and specifically fetal fat segmentation, is known to be a very challenging task. Previous works suggest that segmentation observer variability can be used as a surrogate to estimate the expected accuracy of automatic segmentation \cite{joskowicz2019observervar}. Here we show that fetal subcutaneous fat delineation has a high inter-observer variability with a Dice of 0.738 vs. 0.85-0.95 typically observed for other fetal structures \cite{torrents2019fetalsegmentationreview}, which may be correlated to the difficulty of automatic segmentation. Estrada et al. \cite{estrada2020fatsegnet} report an inter-observer variability of 0.982 and 0.788 Dice scores for adult subcutaneous and visceral fat, respectively. Adult visceral fat segmentation problem is similar to that of fetal subcutaneous fat, as it is a complex, sparse structure with many connected components, due to the intestines and the large surface it covers.

To address these issues, we used a semi-automatic method that reduced the observer variability and shorten segmentation times. These results suggest the advantage of using a semi-automatic segmentation to more efficiently produce more accurate ground-truth annotations used to train an automatic DL method. This scheme is similar to the one described in Kway et al. \cite{kway2021automated} for pediatric subcutaneous and visceral fat segmentations. 


In this work, we explore the use of three neural networks for fetal fat segmentation. In particular, we compare the state-of-the-art models Residual 3D UNet, nn-UNet \cite{isensee2021nnunet} and SWIN-UNetR \cite{tang2022swin}. In previous works \cite{estrada2020fatsegnet,kway2021automated}, variants of 3D-UNet were used to automate fat segmentation. However, nn-UNet and SWIN-UNetR, which yield state-of-the-art results in a variety of biomedical image segmentation scenarios were not evaluated on fat segmentation tasks. In our study, SWIN-UNetR and 3D-UNet achieved similar results, while nn-UNet performed poorly. A possible explanation might relate to the different characteristics of fat segmentation, which differ from what nn-UNet was designed for \cite{isensee2021nnunet}. Related fat segmentation methods achieved a Dice of 0.850 and 0.872 for pediatric and adults, respectively \cite{estrada2020fatsegnet,kway2021automated}. Our result of 0.862 is comparable to those results, and is better than observer variability of manual delineation. Moreover, the automatic method achieved an RVD of 7.11$\%$, which is lower than the interobserver variability RVD of 29.91$\%$ (manual) and 9.26$\%$ (semi-automatic).

Our study have several limitations. First, this is a single center cohort, which may result in poor generalization. Second, we include fetuses with a gestation age of 31 weeks or more, which may limit the automatic segmentation for younger fetuses. However, previous studies showed that the fetal fat is apparent in MRI only from the $28^{th}$ week with rapid third trimester lipid accumulation \cite{blondiaux2018developmental,giza2021water}. Therefore, the pivotal age for fat quantification is similar to that of our study. Third, our cohort size of 57 fetuses is relatively small. Future studies should use a larger, and wider gestational age range to develop and assess the applicability of automatic methods for fetal fat quantification. Lastly, we explored the naive use of state-of-the-art DL methods. Future studies may design a more targeted solution of DL method in order to further improve fetal fat automatic segmentation and reduce the need for manual correction.

\section{Conclusion}
Here we propose the first method to automate subcutaneous fetal fat segmentation. We show that using state-of-the-art segmentation methods and short acquisition time MRI sequences, whole fetal body subcutaneous lipids can be quantified. We anticipate that our method can be used to study normal changes of fetal fat with gestational age and its relation to various abnormal conditions, e.g., gestational diabetes and fetal growth restriction. 

\textbf{Acknowledgements} This research was supported by Kamin Grants [63418, 72126] from the Israel Innovation Authority.

%
%
%
%
\bibliographystyle{splncs04}
\bibliography{allref.bib}

\end{document}